\documentclass[journal]{IEEEtran}
\usepackage{amsmath,amsfonts}
\usepackage{algorithmic}
\usepackage{algorithm}
\usepackage{array}
\usepackage[caption=false,font=normalsize,labelfont=sf,textfont=sf]{subfig}
\usepackage{textcomp}
\usepackage{stfloats}
\usepackage{url}
\usepackage{verbatim}
\usepackage{graphicx}
\usepackage[normalem]{ulem}
\useunder{\uline}{\ul}{}
\usepackage{cite}
\usepackage{amssymb}
\usepackage{xcolor,colortbl}
\usepackage{authblk}
\definecolor{LightCyan}{rgb}{0.88,1,1}
\definecolor{LightOrange}{rgb}{1,1,0.88}
\begin{document}

\title{FusionRF: High-Fidelity Satellite Neural Radiance Fields from Multispectral and Panchromatic Acquisitions}

\author{
        \IEEEauthorblockA{Michael Sprintson, Rama Chellappa, Cheng Peng}\\
       	\vspace{-0.9em}
        \IEEEauthorblockN{\textit{Johns Hopkins University, Baltimore, Maryland, USA}}\\
    }



\maketitle

\begin{abstract}
We introduce FusionRF, a novel framework for digital surface reconstruction from satellite multispectral and panchromatic images. Current work has demonstrated the increased accuracy of neural photogrammetry for surface reconstruction from optical satellite images compared to algorithmic methods. Common satellites produce both a panchromatic and multispectral image, which contain high spatial and spectral information respectively. Current neural reconstruction methods require multispectral images to be upsampled with a pansharpening method using the spatial data in the panchromatic image. However, these methods may introduce biases and hallucinations due to domain gaps. FusionRF introduces joint image fusion during optimization through a novel cross-resolution kernel that learns to resolve spatial resolution loss present in multispectral images. As input, FusionRF accepts the original multispectral and panchromatic data, eliminating the need for image preprocessing. FusionRF also leverages multimodal appearance embeddings that encode the image characteristics of each modality and view within a uniform representation. By optimizing on both modalities, FusionRF learns to fuse image modalities while performing reconstruction tasks and eliminates the need for a pansharpening preprocessing step. We evaluate our method on multispectral and panchromatic satellite images from the WorldView-3 satellite in various locations, and show that FusionRF provides an average of 17\% reduction in depth reconstruction error, and renders sharp training and novel views.
\end{abstract}


\section{Introduction}

\begin{figure}[t!]
    \centering
    \includegraphics[width=1\linewidth]{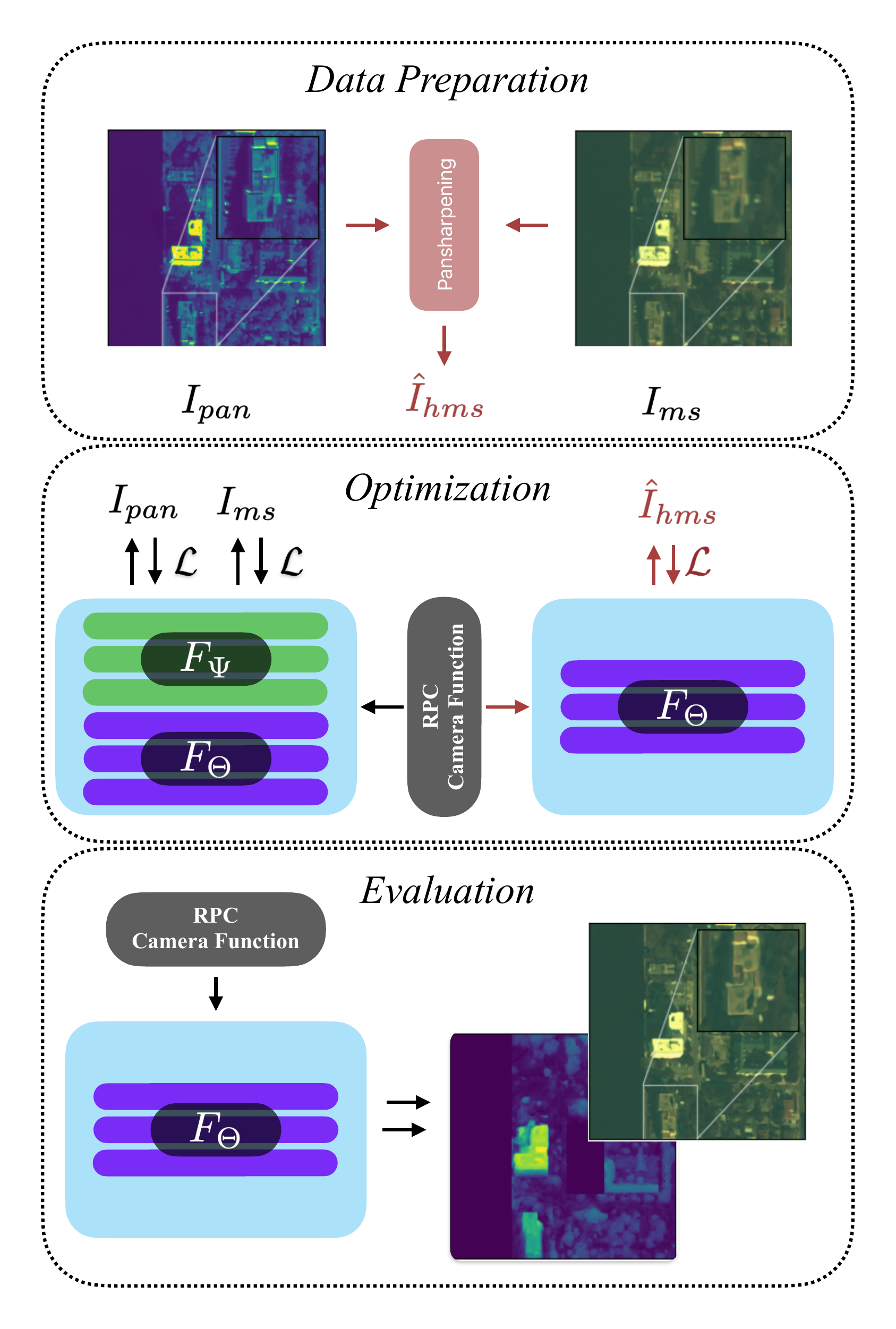}
    \caption{Model Comparison:  
     The optimization of previous models, such as S-NeRF and Sat-NeRF \cite{derksen2021shadow, mari2022sat} are shown in red, first pansharpening the input data and then computing loss only against the resulting pansharpened image. Our method independently trains a NeRF $(F_{\Theta})$ on multispectral and panchromatic images, optimizing against the original input images. The inclusion of the cross-resolution kernel ($F_{\Psi}$) encourages the model to learn to perform image fusion during training. During evaluation, the cross-resolution kernel is disabled, rendering novel view multispectral images with increased spatial resolution. }
    \label{fig:modeldiag}
\end{figure}

\IEEEPARstart{D}{IGITAL} Surface Models (DSMs) allow us to visualize and understand the topology of our world. Historically, elevation data used to construct DSMs has been collected with complex and expensive specialized sensors and satellites such as SAR~\cite{gens1996review} and LiDAR~\cite{collis1970lidar}. Recent work has attempted to reconstruct accurate surface models and DSMs from much more commonly available optical satellite imagery, collected from various angles and over the span of years or months. By generating accurate surface reconstructions, these methods can visualize the scene from novel view angles even under different weather conditions. 

While algorithmic methods~\cite{beyer2018ames, de2014automatic} depend on calibrated stereo or tri-stereo image products, neural rendering methods~\cite{derksen2021shadow, mari2022sat, qu2023sat, mari2023multi} leverage larger datasets of commercially available satellite images. These methods, known as Neural Radiance Fields (NeRF), perform scene reconstruction and surface modeling by optimizing multilayer perceptrons (MLPs) to represent the density and radiance information of a scene, creating an implicit understanding of the scene geometry. The MLPs can then be sampled from various input views for inference, generating depth maps and novel views of the scene. NeRFs have been shown to outperform previous algorithmic methods in surface reconstruction while introducing the ability to recognize and remove transient objects~\cite{mari2022sat}, model the scene under various sun angles~\cite{derksen2021shadow}, and estimate bundle adjustment parameters~\cite{mari2023multi}. Additionally, NeRFs have the ability to model photometric differences between views~\cite{martin2021nerf} and resolve image quality issues such as camera blur and loss of resolution~\cite{ma2022deblur, wang2023bad, peng2023pdrf}.

However, optical data undergoes a significant amount of pre-processing before it is used by a photogrammetry algorithm. Image sensors within satellites simultaneously generate both a single-band panchromatic image with high spatial fidelity and a multispectral image with many narrow wavelength bands. Because collected light in each multispectral band is significantly restricted, the spatial fidelity is substantially reduced. Pansharpening algorithms are employed to perform image fusion, leveraging the spatial resolution in the panchromatic image to upsample the multispectral image, aiming to access the full spatial and spectral information in the satellite images. 

Current NeRF architectures for satellite reconstruction~\cite{qu2023sat, mari2023multi} depend on pansharpening prior to optimization to create training imagery. Pansharpening methods can produce variance in upsampled images and often lack generalizability across satellite image sensors and unseen domains~\cite{deng2022machine}. Since a satellite neural reconstruction model uses the upsampled images as ground truth during optimization, any hallucinations will also be present in the final reconstruction and affect the accuracy of the Digital Surface Model.

We present FusionRF, a neural reconstruction methodology that enables the fusion of panchromatic and multispectral images during reconstruction. While pansharpening methods frame the fusion problem as an upsampling operation to produce one hybrid image, we can process both modalities as inputs of varying spatial and spectral resolution within our reconstruction algorithm. A comparison between methods is shown in Figure \ref{fig:modeldiag}. This approach learns how to fuse information between different modalities on a per-scene basis, removing the risk of pre-trained pansharpening models introducing biases and hallucinations in unseen domains.

Our proposed method, FusionRF, is a neural radiance field which optimizes directly on the full-channel multispectral and panchromatic images from satellite acquisitions. During optimization, fusion is facilitated by a cross-resolution kernel that simulates the loss of resolution in multispectral images, while a modal appearance embedding encodes the characteristics of panchromatic and multispectral views. Image appearance embeddings are employed to resolve photometric differences between images, while transient embeddings in conjunction with uncertainty learning help separate transient phenomena such as cars, construction activity, and seasonal foliage from static objects in the scene. FusionRF learns to perform image fusion during training, producing high-fidelity novel view synthesized images. During evaluation, the cross-resolution kernel is disabled, allowing the NeRF to produce high-resolution novel views which are consistent with the input scene and outperform publicly available state-of-the-art methods in 3D reconstruction tasks.

Our main contributions can be summarized as follows: 

% \vspace{-0.1em}

\begin{itemize}{
\item A novel satellite NeRF with an embedding strategy that enables the use of full-channel multispectral and panchromatic inputs, removing the dependence on satellite image pre-processing.
\item A sparse cross-resolution kernel which models the resolution loss in multispectral images, allowing for intrinsic pansharpening within the optimization process.
\item A comprehensive set of experiments that demonstrate an average 17\% decrease in depth reconstruction error, highlighting the value of removing the pansharpening stage from satellite reconstruction algorithms.
}
\end{itemize}

\section{Background}
\subsection{Pansharpening}

Commercial satellites are typically equipped with a pair of sensors that separately gather detailed spatial and spectral data. Some applications, such as water resource management~\cite{wood2011hyperresolution} and land use management~\cite{jarnstedt2012forest, nurminen2013performance, bandara2011automated} require gathered light to be split into narrow frequency bands to analyze and monitor water quality and vegetation health. Multispectral sensors filter incoming light into these specialized channels and require larger photosites, trading spatial fidelity for spectral resolution. Panchromatic sensors accept one broad band of light, creating a single band image with high spatial resolution.

For compatibility with contemporary image analysis algorithms and reconstruction models, the panchromatic image $I_{pan}$ and multispectral image $I_{ms}$ are often fused into one hybrid image $I_{hms}$. This approach, known as pansharpening, attempts to exploit the spatial information in $I_{pan}$ as a guide for upsampling $I_{ms}$ into $I_{hms}$. Many such pansharpening methods exist, from algorithmic approximations of the human optical cortex's understanding of sharpness and color~\cite{palsson2013new, kang2013pansharpening, padwick2010worldview} to deep learning methods~\cite{masi2016pansharpening, wei2017multi, deng2020detail, wang2024towards}. 

However, while deep learning-based methods are more effective than algorithmic methods, they are dependent on large datasets, are susceptible to poor generalization, require large training times~\cite{cao2024zero}, and often must be trained on downsampled data. Because no ground truth $I_{hms}$ with high spatial and spectral resolution exists, deep learning models treat the original $I_{ms}$ as ground truth and create a downsampled $I_{lms}$ as the new model input. In training, the models upsample $I_{lms}$ to $I_{ms}$ and depend on scale-invariance to upsample $I_{ms}$ to $I_{hms}$ in inference, which is not guaranteed in practice~\cite{deng2022machine}. More recent work focuses on unsupervised learning methods~\cite{qu2020unsupervised, luo2020pansharpening, ma2020pan, zhou2021pgman, cao2024zero} which operate on full resolution images. These methods are highly sensitive to the complex loss function chosen for comparing images, require a very large amount of training data, and suffer from poor generalization to unseen domains~\cite{zhang2023panchromatic}. As satellite reconstruction models use only a few views of one area,
the perceptual errors resulting from the domain shift of the pansharpening model are amplified in the final reconstruction.

\subsection{Neural Radiance Fields and Variants}

Neural Radiance Field~\cite{mildenhall2021nerf} methods attempt to encode a scene within a fully connected neural network $F_{\Theta}$: 

\vspace{-0.8em}
\begin{equation}
\label{fthetaeq}
F_{\Theta}(\gamma_{x}(\textbf{x}), \gamma_{d}(\textbf{d})) = (\sigma, \textbf{c}).
\end{equation}

Here, $\gamma_x$ and $\gamma_d$ represent positional encodings that convert camera coordinates $\textbf{x}$ and camera direction $\textbf{d}$ to a higher dimensional representation. The function $F_{\Theta}$ is able to represent the scene as a combination of volume density $\sigma$ and predicted radiance $\textbf{c}$, which allows a viewer to render the scene from a novel input camera position and direction. A ray $\textbf{r}(t) = \textbf{o} + t\textbf{d}$ with origin $\textbf{o}$ and direction $\textbf{d}$ is projected from each pixel in the generated image towards the scene, and classical volume rendering is used to render the predicted color $\hat{\textbf{C}}(\textbf{r})$ by sampling $N$ locations along the ray:

\vspace{-1.5em}
\begin{equation}
\label{creq}
\hat{\textbf{C}}(\textbf{r}) = \sum_{i=1}^{N} T_i(1-e^{-\sigma_{i}\delta_{i}}) \textbf{c}_{i} \;\; \text{where}  \;\; T_i=e^{-\sum_{j=1}^{i-1} \sigma_j \delta_j}.
\end{equation}

Here, $T_i$ represents the transmittance, or the probability that the ray travels without interference, while $\delta_i$ represents the distance between samples. The predicted color $\hat{\textbf{C}}(\textbf{r})$ is weighted by the transmittance and opacity of the scene. This allows the color to be controlled by the predicted scene content along the entire ray, taking into account occlusions and reflections. The loss $\mathcal{L}$ is computed as the error between the colors generated by the rendered rays and the ground truth pixel color $\textbf{C}(\textbf{r})$:
$$\mathcal{L} = \vert\vert \hat{\textbf{C}}(\textbf{r}) - \textbf{C}(\textbf{r}) \vert\vert^{2}.$$

\vspace{-1.5em}
\subsection{Resolution Recovery NeRF Variants}

Recent methods~\cite{ma2022deblur, wang2023bad, peng2023pdrf} address the problem of deblurring input images by modeling a blurry image as a convolution of a blur kernel with a sharp image. For a particular pixel $p$ with color $\textbf{c}_p$, the blurry color can be calculated as a convolution $\textbf{b}_p = \textbf{c}_p \circledast \textbf{H}$. The blur kernel $\textbf{H}$ typically consists of a $K \times K$ matrix. However, while a typical deblurring model convolves a predicted image with $\textbf{H}$ to compare to the blurry ground truth, a NeRF model cannot due to the computational complexity of rendering $K^2 \times h \times w$ rays for an image with dimensions $(h,w)$. These methods seek to resolve this issue by approximating the dense blur kernel $\textbf{H}$. A small set of points are selected and rays $\{\textbf{q}\}$ are generated as a sparse approximation. A separate MLP then predicts the weights $w_\textbf{q}$, as well as regresses the optimal positions of the origins and destinations of $\textbf{q}$. Each ray $\textbf{q}$ generates a color $\textbf{c}_q$ which is then weighted by $w_\textbf{q}$ to produce the blurry color: $\textbf{b}_p = \sum_{\textbf{q}} \textbf{c}_\textbf{q} w_\textbf{q}$. The deformation of rays allows for the kernel to model complex motion and defocus blur. Because the underlying NeRF model predicts sharp images which are then deformed by the blur kernel to match the blurry inputs, we can remove the kernel in evaluation to render sharp images of the scene.

\subsection{Satellite NeRF Models}

While NeRF depends on a camera position $\textbf{x} = (x,y,z)$, images from observational satellites instead provide Rational Polynomial Coefficients (RPC). These coefficients define a geometric correction projection function for georeferencing pixels in the satellite image. While prior works such as \mbox{S-NeRF}~\cite{derksen2021shadow} first estimate and replace the RPC sensor model with an approximate pinhole model, \mbox{Sat-NeRF}~\cite{mari2022sat} samples the rays directly from the RPC sensor model. In this approach, the bounds of each ray become the maximum and minimum altitudes of the scene $h_{max}$ and $h_{min}$. Rays are then sampled by georeferencing pixel values at the minimum and maximum altitudes, sampling intermediate points during training. Both \mbox{S-NeRF} and \mbox{Sat-NeRF} rely on a shadow-aware irradiance model to render the effects of shadows, while Sat-NeRF additionally incorporates an uncertainty-learning approach similar to NeRF-W to effectively mask some of the transient objects in the scene.

EO-NeRF~\cite{qu2023sat} expands on Sat-NeRF by rendering shadows using scene geometry and solar direction instead of predicting them with an MLP head. Additionally, the model incorporates bundle adjustment within the training optimization instead of as a separate pipeline and allows the use of uncorrected satellite imagery.  RS-NeRF~\cite{xie2023remote} adds hash encoding to speed up optimization, while SparseSat-NeRF~\cite{zhang2023sparsesatnerfdensedepthsupervised} utilizes depth supervision to enhance performance.

The model proposed by Pic \textit{et al.}~\cite{10641439} also optimizes on both multispectral and panchromatic images, relying on a linear function to convert between the R, G, and B subset of bands within a multispectral and the panchromatic band. This method uses both a loss against the predicted RGB image and a loss between the converted panchromatic prediction and the original panchromatic image. In comparison, our method directly predicts the panchromatic and multispectral images, performing fusion implicitly within the NeRF. This allows our model direct access to the spatial resolution found in the panchromatic image instead of through a limited learned function. Unlike~\cite{10641439}, a specific relationship between multispectral and panchromatic images is not enforced, uniquely allowing our model to naturally handle datasets with differing numbers of multispectral images and panchromatic images.

\section{Method}

For each input satellite view, both a multispectral image $I_{ms}$ with dimensions $(h,w)$ and panchromatic image $I_{pan}$ with dimensions $(rh,rw)$ are provided, with $r$ representing the scaling factor between the two images. Previous approaches require a pansharpening function $P$ to create an image \mbox{$I_{ps} = P(I_{ms}, I_{pan})$} and use this image as input to the NeRF model, while our approach only requires a bilinear interpolation function to create $I_{lms} = \uparrow_{r}(I_{ms})$. Since both images are of the same resolution and cover the same geographic area, the pixels in each image are aligned. We model the blurring process as a convolution $I_{lms} = I_{hms} \circledast \textbf{H}$, where $I_{hms}$ represents a predicted multispectral image with the same spatial resolution as $I_{pan}$. Here, \textbf{H} represents a blurring kernel that causes resolution loss. It is important to note that no ground truth image $I_{hms}$ exists with high spectral and spatial resolution image to supervise the model.

\begin{figure*}[t!]
    \centering
    \includegraphics[width=\linewidth]{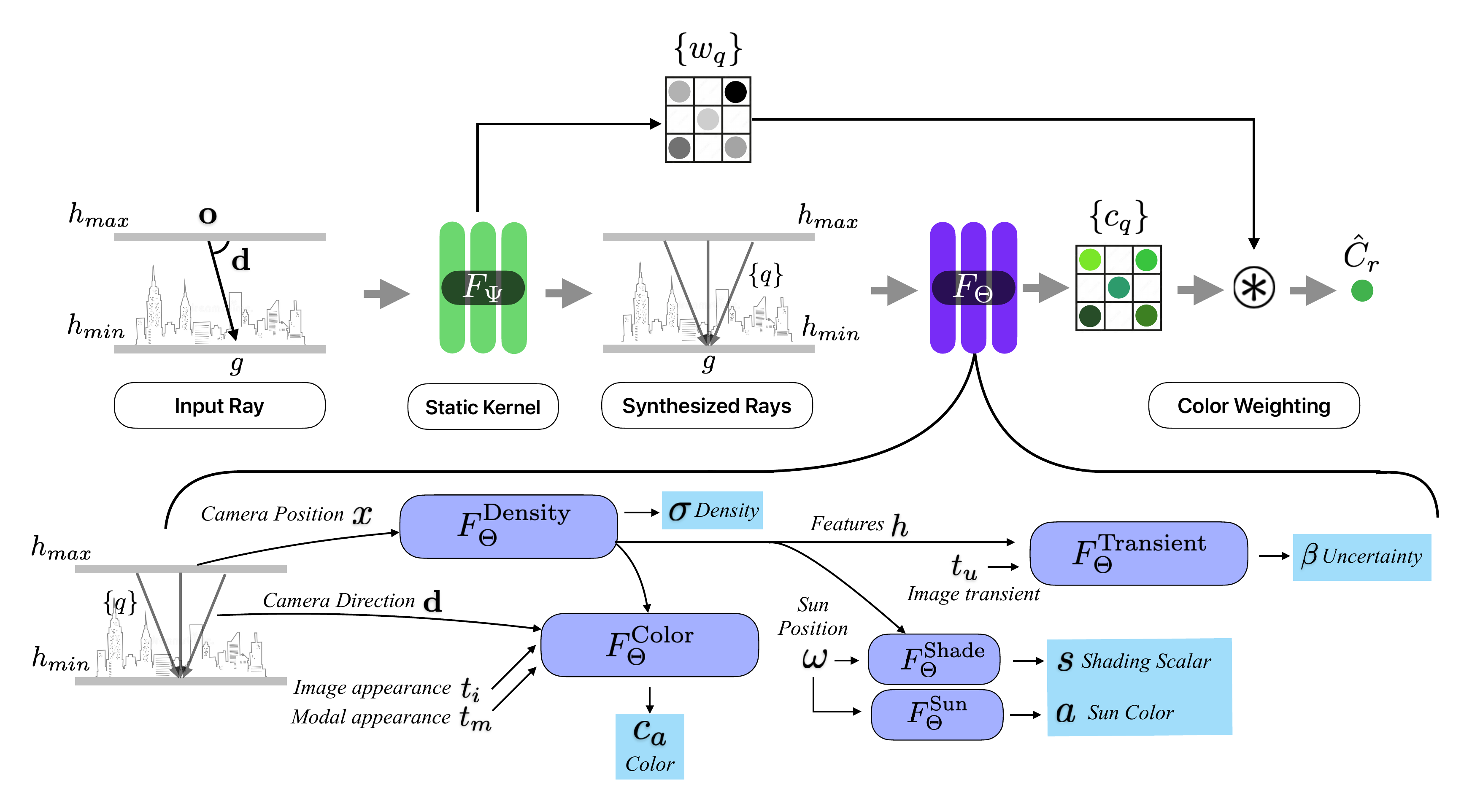}
    \caption{Our Network Architecture: Every input ray $r$ is projected from origin $\textbf{o}$ in $I_{lms}$ and $I_{pan}$ to the same ground point $g$. The sparse cross-resolution kernel $F_{\Psi}$ predicts weights $w_q$ for static locations $\{q\}$ surrounding $\textbf{o}$, which are then combined with the color predictions $\{c_q\}$ to produce the final output color. }
    % \vspace{-1em}
    \label{fig:netarch}
\end{figure*}
\subsection{RPC-based Sampling}

Similarly to Sat-NeRF~\cite{mari2022sat}, the origin $\textbf{o}$ of each ray is calculated by georeferencing the origin pixel $(x, y)$ with $h_{max}$. The destination $\textbf{g}$ of each ray can be calculated similarly by georeferencing $(x, y)$ with $h_{min}$. The origin $\textbf{o}$ and destination $\textbf{g}$ are converted to a geocentric coordinate system, and the direction $\textbf{d}$ of each ray is calculated as: 

\begin{equation}
\label{direq}
\textbf{d} = \frac{\textbf{g} - \textbf{o}}{\vert\vert \textbf{g} - \textbf{o} \vert\vert^2}.
\end{equation}

The ray parameters $\textbf{o}$ and $\textbf{d}$ are then normalized to reduce the required precision for the model. Each ray can then be sampled between 0 and $\vert\vert \textbf{g} - \textbf{o} \vert\vert^2$. 

\subsection{Cross-Resolution Kernel}
\label{subsec:crk}

We introduce a novel cross-resolution kernel $F_{\Psi}$ to model the resolution difference between image $I_{lms}$ and image $I_{pan}$. Optimizing a NeRF over $I_{lms}$ would require that the model predict the same low resolution $I_{lms}$ for each ray originating in the image. As we wish to prevent our model from learning directly from $I_{lms}$, we estimate a relationship $I_{lms} = (I_{pred} \circledast F_{\Psi})$ where $I_{pred}$ represents the image predicted by the NeRF $F_{\Theta}$. Since the datasets for satellite reconstruction tasks contain relatively few images, we design a kernel $F_{\Psi}$ with minimal parameters and a fixed sparse support of nine rays in a $2r{+}1 \times 2r{+}1$ window. This pattern was chosen to align with the upscaling factor $r$ of the $I_{lms}$ image, placing each support ray $r$ pixels away in each dimension from all other support rays. This sampling strategy ensures that support rays avoid redundant information from neighboring pixels in $I_{lms}$ caused by upscaling, while capturing information from neighboring pixels in the original $I_{ms}$ image.

For a pixel $p$, scaling factor $r$, and coordinates $(x,y)$, we define the fixed positions:

\vspace{-0.5em}
\begin{equation}
\label{qeq}
\{ q \} = \{(x+j,y+k) \mid j,k \in \{-r,0,r\}\}.
\end{equation}
Our $F_{\Psi}$ MLP predicts the weights:

\vspace{-0.5em}
\begin{equation}
\label{qweq}
w_\textbf{q} = \textbf{$F_{\Psi}$}(p,q)
\end{equation}

for each ray $\textbf{q}$, which are combined in a similar way to DeBlur-NeRF~\cite{ma2022deblur}. The predicted color $\hat{\textbf{C}}({\textbf{r}})$ is computed as the weighted sum of the color from the sparse support rays, using weights from $F_{\Psi}$:
\vspace{-0.5em}
\begin{equation}
\label{chatreq}
\hat{\textbf{C}}({\textbf{r}}) = \sum_{\{\textbf{q}\}} w_\textbf{q} \textbf{c}_{\textbf{q}}.
\end{equation}

Empirically, we find that the inclusion of $F_{\Psi}$ in this fashion allows the multispectral images rendered without a sparse cross-resolution kernel to be considerably sharper. A more complex design of $F_{\Psi}$, such as randomizing the positions or predicting deformations of ${\textbf{q}}$, leads to an over-parameterized model and negatively impacts reconstruction accuracy. 

However, the addition of this kernel alone does not render sharp novel view images. Without the presence of panchromatic imagery and intrinsic pansharpening as described in Section \ref{sec:ip}, the model does not have sufficient information to recreate ground-level details in multispectral imagery. This limitation is demonstrated in the experiments in Section \ref{sec:exp}.

\subsection{Intrinsic Pansharpening}
\label{sec:ip}

% \begin{figure}[!t]
% \centering
% \subfloat[]{\includegraphics[width=2.04in]{figures/modela.png}%
% \label{fig_first_case}}
% \hfil
% \subfloat[]{\includegraphics[width=1.4in]{figures/modelb.png}%
% \label{fig_second_case}}
% \caption{Model Diagram:  
%     (\ref{fig_first_case}) Our method independently trains a NeRF $(F_{\Theta})$ on multispectral and panchromatic images, computing loss against the original input images. The inclusion of the sharpening kernel ($F_{\Psi}$) encourages the model to learn to perform image fusion during training. (\ref{fig_second_case}) During evaluation, the sharpening kernel is disabled, rendering novel view multispectral images with increased spatial resolution.}
% \label{fig_sim}
% \end{figure}

\begin{figure*}[!t]
    \centering
    \includegraphics[width=1\linewidth]{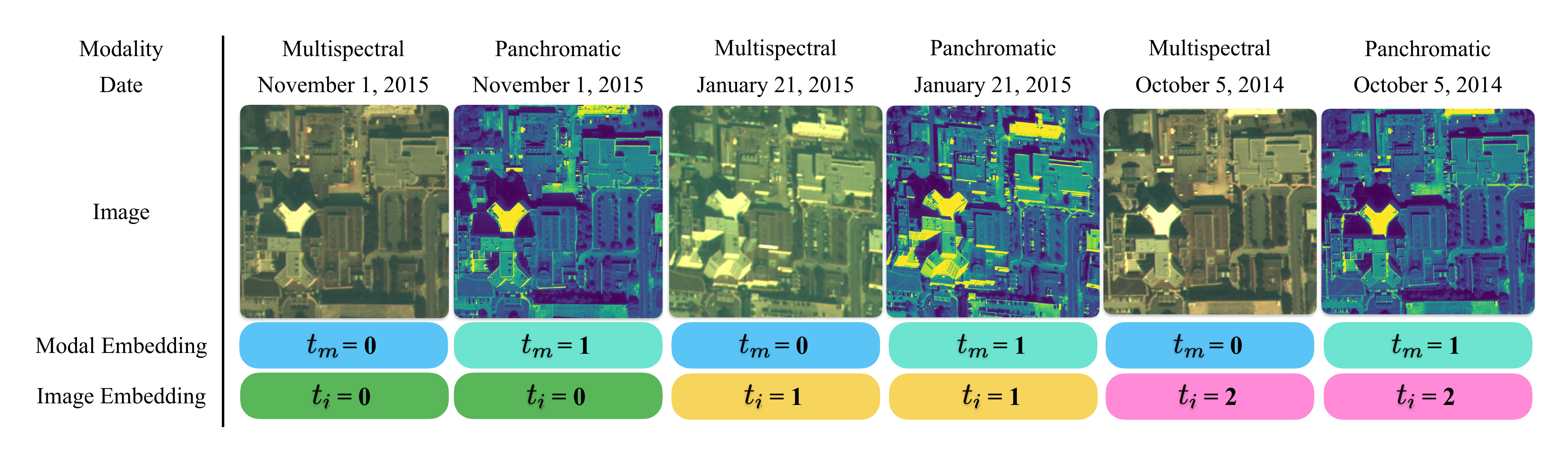}
    \caption{Embedding Diagram:  
    One modal embedding is shared across all panchromatic images and another across all multispectral images, allowing the model to encode modal information in a uniform representation. Each image view, represented by date of capture, also shares one image embedding across both the panchromatic and multispectral images. In the case of a view containing only one modality, the image embedding is unique to that image. }
    \label{fig:embdiag}
\end{figure*}

The images $I_{lms}$ and $I_{pan}$ are collected simultaneously and aligned during collection. Therefore, a pixel $p$ at location $(x,y)$ describes the same geographic location $\textbf{g}$ in both images. FusionRF takes both images as input independently of each other, creating two rays $\textbf{r}_{ms}$ and $\textbf{r}_{pan}$ which both begin at $(x,y)$ and describe $\textbf{g}$. These rays differ in their predicted colors $\hat{\textbf{C}}(\textbf{r})$: $\textbf{r}_{ms}$ predicts the color of $p$ in $I_{lms}$ and $\textbf{r}_{pan}$ in $I_{pan}$. Therefore, the loss computation for each image will be independent, but will each be calculated against the same $\textbf{g}$. For the multispectral image, the loss $\mathcal{L}_{Image} = \vert\vert \hat{\textbf{C}}({\textbf{r}}) - \textbf{C}_{lms}({\textbf{r}}) \vert\vert^{2}$ encourages the model to predict a $\hat{\textbf{C}}(\textbf{r})$ with a blurry color and good multispectral resolution across the full eight output channels. In contrast, the loss for the panchromatic image \mbox{$\mathcal{L}_{Image} = \vert\vert \hat{\textbf{C}}({\textbf{r}}) - \textbf{C}_{pan}({\textbf{r}}) \vert\vert^{2}$} encourages the model to predict a sharp image with panchromatic color. Here, $\textbf{C}_{lms}({\textbf{r}})$ and $\textbf{C}_{pan}({\textbf{r}})$ represent the ground truth color for the pixel $p$ in the input images $I_{lms}$ and $I_{pan}$. 

We introduce a model embedding $t_m$ for both $F_{\Psi}$ and $F_{\Theta}$ to indicate the image type. This embedding is shared across each modality, providing the model with a small shared vector in which to encode the properties of that modality. We believe that this encourages more effective fusion of the information in each modality into a single model $F_{\Theta}$. 

We also incorporate an image-specific appearance embedding $t_i$ as an input to the color-producing section of the network $F_{\Theta}^{\text{Color}}$, similarly to NeRF-W~\cite{martin2021nerf}. This prevents the appearance embedding from affecting the volume density prediction of the network, while allowing the model to remember variations in color between multispectral images. 

Since panchromatic and multispectral images are geometrically aligned and captured under identical weather and lighting conditions, we enforce that the embedding $t_i$ is shared between both $I_{lms}$ and $I_{pan}$. This requirement ensures that only the modal embedding $t_m$ is used to separate images from the two modalities. This embedding strategy is visualized in Fig.~\ref{fig:embdiag}.

It is important to note that FusionRF can still operate with limited panchromatic imagery. While a traditional pansharpening method depends on the panchromatic pair $I_{pan}$ for a multispectral image $I_{lms}$ to perform sharpening, our model is able to share information between images from the same scene through the $F_{\Theta}$ MLP to fill in for a missing $I_{pan}$ pair. Information from the other available $I_{pan}$ images in the scene can still be applied to improve the sharpness of the rendered view of $I_{lms}$. Therefore, intrinsic pansharpening within the training dataset is still possible with limited panchromatic data and can even be done on novel multispectral views of the scene. 

Our approach, shown graphically in Figure \ref{fig:netarch}, encourages the model to develop images that exactly match the original $I_{lms}$ and $I_{pan}$, removing the requirements for hand-crafted image fusion metrics or artificial downsampling of training data. Rather than a separate pre-processing operation, pansharpening becomes an internal task for our model to solve in the production of high quality novel views. The model's primary task is novel view synthesis, which ensures the predictions are grounded in the observed scene, reducing the likelihood of hallucinations. Since only one scene is considered in the training stage, the model cannot carry over artifacts from the domain of other scenes. 

\subsection{Multispectral NeRF}

Multispectral satellite images provide wavelength information in many small ranges. In commonly available imagery from commercial satellites, this can range from 4-12 bands~\cite{williams2006landsat, cantrell2021system, phiri2020sentinel}. Previous approaches~\cite{derksen2021shadow, mari2022sat, mari2023multi, 10641439} have required that the spectral bands be filtered to three: those that accept the primary colors red, green, and blue. This means that the underlying NeRF only has access to a limited portion of the overall spectral band. Our model aims to share information between the panchromatic and multispectral bands and therefore does not restrict the channel capacity of the $F_{\Theta}$ MLP~\cite{mildenhall2021nerf}. For panchromatic images, we replicate the information in the panchromatic image across all channels. 

Our model's MLP consists of two stages. First, $F_{\Theta}^{\text{Density}}$ predicts the volume density $\sigma$ as a function of the camera position $\textbf{x}$:
\vspace{-0.2em}
\begin{equation}
\label{fdensity}
F_{\Theta}^{\text{Density}}(\gamma_{x}(\textbf{x})) = (\sigma, \textbf{h}),
\end{equation}

Here, $\gamma_{x}$ represents a positional encoding which converts the camera coordinates to a higher dimensional representation. The MLP $F_{\Theta}^{\text{Color}}$ accepts features $h$ from the first stage as well as self-optimized image embeddings $t_i$ and $t_m$ and view direction $\textbf{d}$ to estimate the initial color $\textbf{c}_{\textbf{a}}$:

\vspace{-0.5em}
\begin{equation}
\label{fcoloreq}
F_{\Theta}^{\text{Color}}(\textbf{h}, \gamma_{d}(\textbf{d}), t_m, t_i) = (\textbf{c}_\textbf{a}).
\end{equation}

To encourage $F_{\Theta}$ to disambiguate between panchromatic and multispectral images in our dataset, we incorporate a modal embedding $t_m$ of size $(2,n)$. For each ray, $t_m$ is indexed to return the specific embedding of size $n$ for the ray's modality. 

Following S-NeRF~\cite{derksen2021shadow}, we adopt two additional components: Layer $F_{\Theta}^{\text{Sun}}(\boldsymbol{\omega}) = \textbf{a}$ predicts the ambient color of shaded areas based solely on the sun position $\boldsymbol{\omega}$ and $F_{\Theta}^{\text{Shade}}(\boldsymbol{\omega}, \textbf{h}) = s$ predicts the impact of shadow on the rendered ray. These outputs are then combined to create a final output color:

\vspace{-0.7em}
\begin{equation}
\label{crweq}
\hat{\textbf{C}}(\textbf{x}, \textbf{d}, \boldsymbol{\omega},t_m,t_i) = \textbf{c}_\textbf{a} (s + \textbf{a}(1-s)).
\end{equation}

\vspace{-1.7em}

\begin{figure*}[b!]
    \centering
\includegraphics[width=1\linewidth]{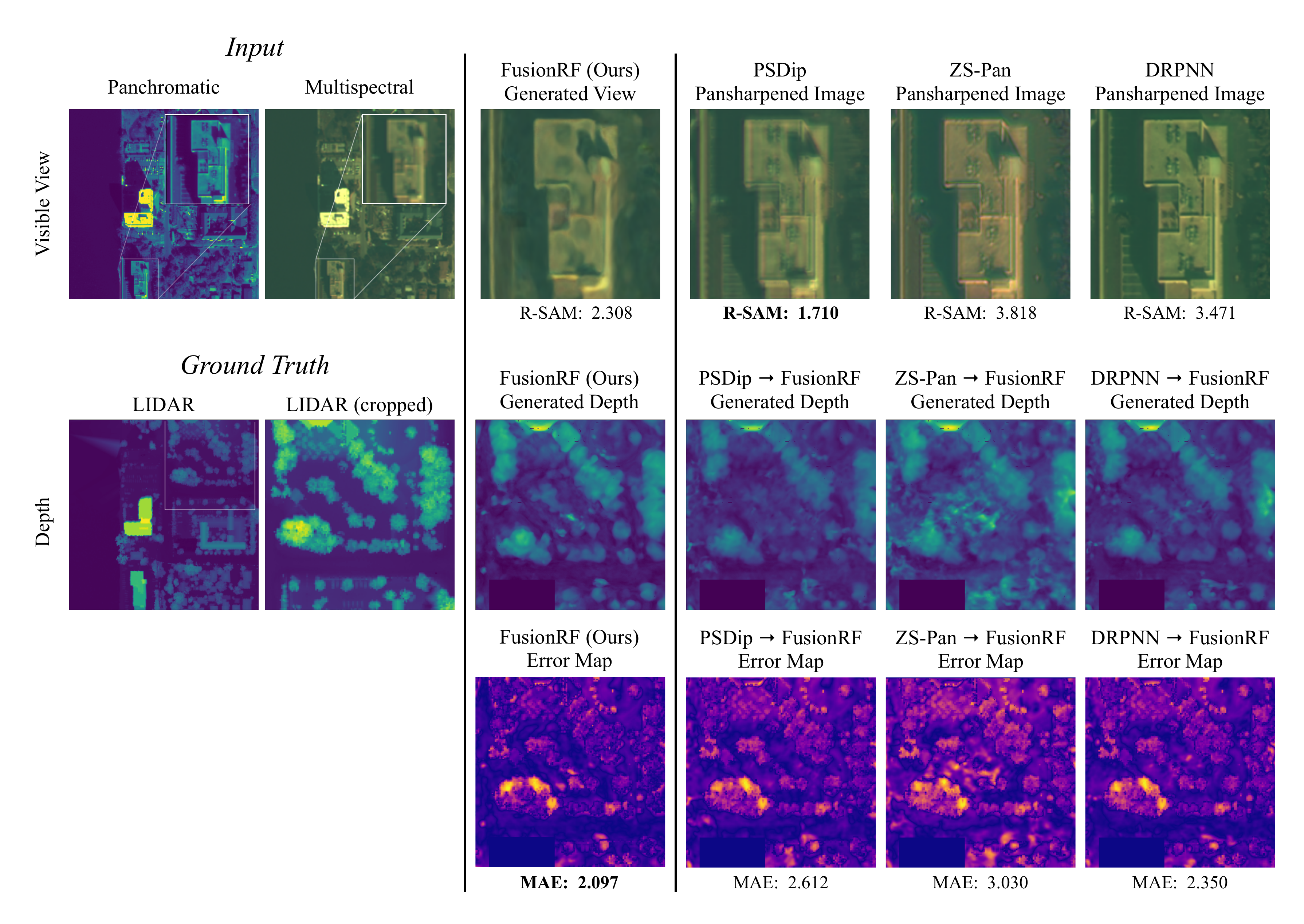}
    \caption{Pansharpening Visual Comparison: The first row shows the original multispectral and panchromatic images alongside FusionRF's generated image and pansharpened images from deep learning methods. The images show the tendency of pansharpening methods to hallucinate color casts and additional details on buildings and shows the variance in the final result produced by pansharpening methods. The second row shows the LIDAR depth map provided by the dataset, along with the rendered depth from a FusionRF model trained on the raw panchromatic and multispectral data along with FusionRF models trained on pansharpened imagery. Finally, the third row shows the error maps of these reconstructions against the LIDAR depth map. This comparison shows that achieving optimal image quality does not improve depth reconstruction.}
    \label{fig:sharppancomp}
\end{figure*}

\subsection{Transient Embedding}

Due to the nature of satellite orbits around the Earth, the collection times for a particular scene can span a long period. While appearance embedding attempts to capture the variations in overall lighting and photometry, a transient embedding is also employed to capture the uncertainty of each pixel in the scene~\cite{martin2021nerf, mari2022sat}. This is particularly useful in our case, as adapting the reconstruction loss to ignore areas of constant change such as parking lots, construction zones, and treetops allows the model to better refine static areas such as buildings. In our model the uncertainty $\beta$ is predicted as a function of previous layer features $h$ and transient embedding $t_u$ through layer $F_{\Theta}^{\text{Transient}}(\textbf{h}, t_u) = \beta$, similarly to NeRF-W~\cite{martin2021nerf} and Sat-NeRF~\cite{mari2022sat}. The uncertainty $\beta$ attenuates the importance of a pixel in the network's loss function:
\vspace{-0.1em}
\begin{equation}
\label{lossfunc}
\mathcal{L} = \frac{\vert\vert \hat{\textbf{C}}({\textbf{r}}) - \textbf{C}_{lms}({\textbf{r}}) \vert\vert^{2}}{2\beta_r^2} + \frac{\log(\beta(\textbf{r}) + \beta_{min}) + \beta_{offset}}{2}.
\end{equation}

While the transient mask allows the model to selectively ignore pixels it believes belong to transient objects, it can also be abused to ignore areas that are simply difficult to reconstruct. As a safeguard, the balancing term $\log(\beta + \beta_{min}) + \beta_{offset}$ is introduced to penalize excessive labeling of uncertain regions to drive down the training loss.

\begin{figure*}[t!]
    \centering
\includegraphics[width=1\linewidth]{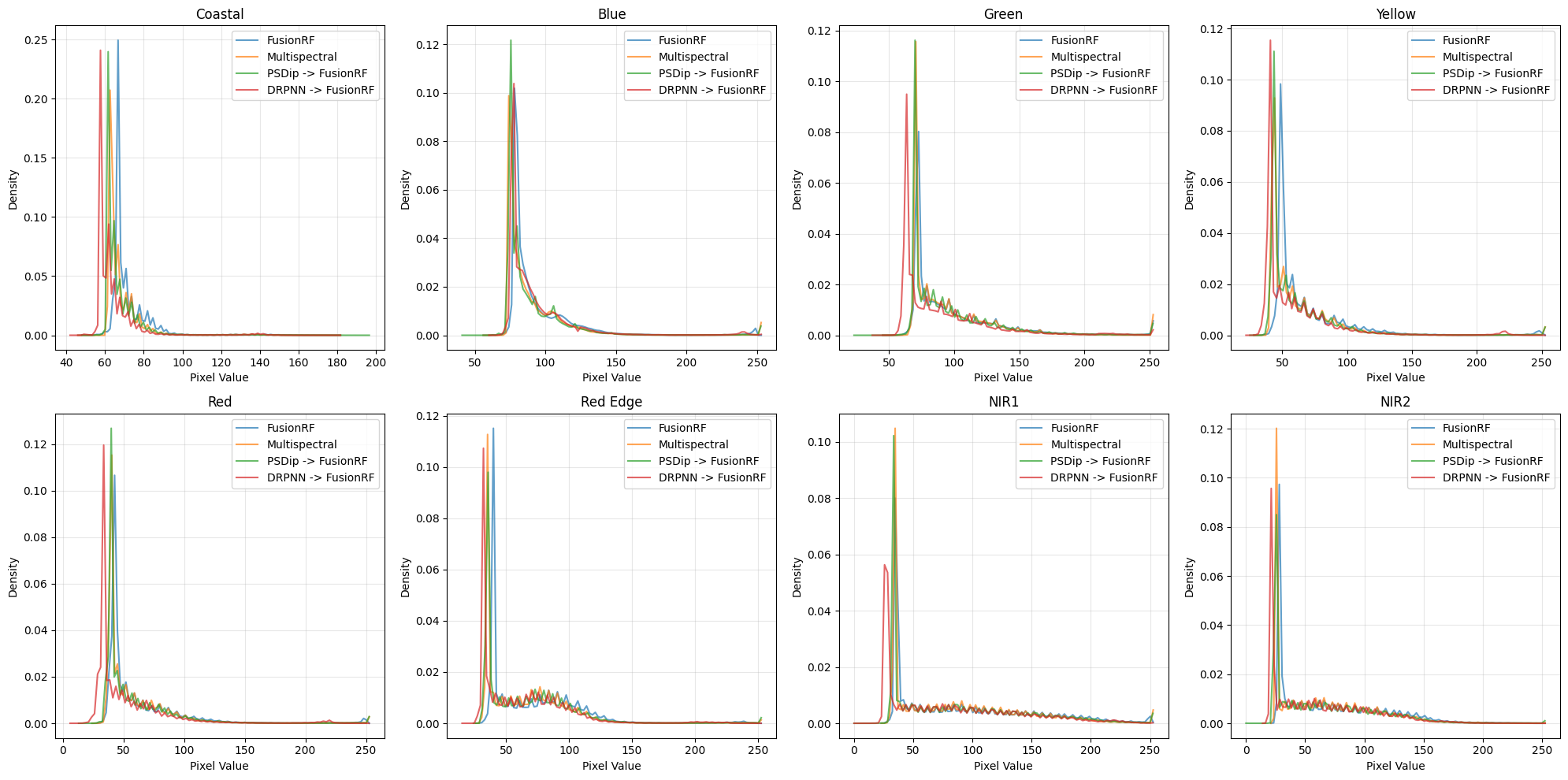}
    \caption{Spectral Comparison: For each of the eight bands of the WorldView-3 image shown in Figure \ref{fig:sharppancomp}, we plot the spectral density in the form of a histogram of pixel values. We display results for the FusionRF generated views from the multispectral and panchromatic data as well as the images pansharpened by PSDip and DRPNN.}
    \label{fig:spectralcomp}
\end{figure*}

\vspace{-0.5em}
\section{Experiments}
\label{sec:exp}

\subsection{Pansharpening Effectiveness Comparison}

We perform two experiments to comparatively evaluate the ability of FusionRF to replace the pansharpening pre-processing stage. First, we measure the depth reconstruction quality under two conditions: FusionRF (with no cross-resolution kernel $F_{\Psi}$) trained on pansharpened images from state-of-the-art deep learning methods, and FusionRF (with $F_{\Psi}$) trained on the original multispectral and panchromatic images. The pansharpening images were created by four state-of-the-art methods: DRPNN~\cite{wei2017boosting}, FusionNet~\cite{deng2020detail}, \mbox{ZS-Pan}~\cite{CAO2024102001}, and \mbox{PSDip}~\cite{10744593}.

\newcolumntype{z}{>{\columncolor{LightCyan}}c}

\begin{table*}[b!]
\caption{Pansharpening Results Comparison: All methods were tested on full-size multispectral and panchromatic data. Best results bolded, second best underlined. Higher values are better for R-Q2n, and lower are better for R-ERGAS, R-SAM, and MAE.}
\label{panresults}
\centering
\centering
\large
\resizebox{\textwidth}{!}{%
\begin{tabular}{c|cccz|cccz|cccz|cccz|}
\cline{2-17}
 & \multicolumn{16}{c|}{\rule{0pt}{2ex}R-Q2n : $\uparrow$  \qquad \qquad R-ERGAS, R-SAM, Depth MAE: $\downarrow$} \\ \cline{2-17} \rule{0pt}{3ex}
 & \multicolumn{4}{c|}{\rule{0pt}{2ex}004} & \multicolumn{4}{c|}{068} & \multicolumn{4}{c|}{214} & \multicolumn{4}{c|}{260} \\ \cline{2-17} \rule{0pt}{3ex}
 & R-Q2n  & R-ERGAS  & R-SAM  & Depth MAE  & R-Q2n & R-ERGAS & R-SAM & Depth MAE & R-Q2n & R-ERGAS & R-SAM & Depth MAE & R-Q2n & R-ERGAS & R-SAM & Depth MAE \\ \hline
\multicolumn{1}{|c|}{\rule{0pt}{3ex}DRPNN} & 0.712 & 10.939 & 3.342 & {\ul 1.700} & 0.793 & 12.780 & 3.179 & \textbf{1.460} & 0.756 & 11.890 & 3.007 & {\ul 2.359} & 0.641 & 12.744 & 3.471 & {\ul 2.350} \\
\multicolumn{1}{|c|}{\rule{0pt}{4ex}FusionNet} & {\ul 0.808} & {\ul 7.500} & {\ul 2.126} & 1.716 & 0.828 & 10.536 & {\ul 2.008} & 1.586 & {\ul 0.782} & 10.082 & {\ul 1.944} & 2.747 & {\ul 0.717} & {\ul 9.906} & {\ul 1.967} & 2.367 \\
\multicolumn{1}{|c|}{\rule{0pt}{4ex}ZS-Pan} & 0.679 & 9.436 & 3.299 & 2.016 & 0.754 & 16.568 & 3.073 & 2.038 & 0.748 & 9.904 & 3.371 & 3.026 & 0.616 & 10.823 & 3.818 & 3.030 \\
\multicolumn{1}{|c|}{\rule{0pt}{4ex}PSDip} & \textbf{0.838} & \textbf{7.171} & \textbf{1.899} & 1.725 & \textbf{0.888} & \textbf{8.279} & \textbf{1.848} & 1.747 & \textbf{0.831} & \textbf{8.271} & \textbf{1.863} & 2.696 & \textbf{0.762} & \textbf{8.643} & \textbf{1.710} & 2.612 \\
\multicolumn{1}{|c|}{\rule{0pt}{4ex}\begin{tabular}[c]{@{}c@{}}FusionRF\\ (Ours)\end{tabular}} & 0.793 & 8.242 & 2.189 & \textbf{1.438} & {\ul 0.857} & {\ul 8.726} & 2.204 & {\ul 1.468} & 0.754 & {\ul 9.498} & 2.375 & \textbf{2.324} & 0.665 & 10.400 & 2.308 & \textbf{2.097} \\ \hline
\end{tabular}%
}
\end{table*}

Second, we evaluate FusionRF's ability to produce effectively fused views of training images using full-resolution indexes R-Q2n, R-ERGAS, and R-SAM for assessing the quality of pansharpening algorithms. The Q2n~\cite{garzelli2009hypercomplex}, ERGAS~\cite{wald2000quality}, and SAM~\cite{kruse1993spectral} indexes were originally designed to assess the performance of reduced-resolution pansharpening methods, but were adapted by Scarpa and Ciotola~\cite{fullresqa} to full resolution indexes through a reprojection protocol to resolve the issue of scale variance in the original multispectral image and the resulting pansharpened image. 

Columns R-Q2n, R-ERGAS, and R-SAM in Table \ref{panresults} show that pansharpening algorithms achieve strong image fusion quality in human perceptual visual quality metrics. However, this performance does not correlate to an improvement in column Depth MAE which represents 3D reconstruction accuracy results as described in Section \ref{drc}. These results show a disconnect between optimization of images for human perception and depth reconstruction. The improvements in image quality index performance that pansharpening algorithms achieve do not necessarily result in an improvement in depth estimation performance. FusionRF trained on panchromatic and multispectral images with $F_{\Psi}$ enabled avoids this pansharpening algorithm selection problem by intrinsically pansharpening the input data, providing the best depth reconstruction performance while still being competitive in pansharpening metrics. Figure \ref{fig:spectralcomp} shows the difference in spectral density between the two best performing pansharpening methods and the FusionRF result.

These results show that the chosen pansharpening algorithm significantly influences the performance of the downstream reconstruction algorithm, and intrinsic pansharpening through the $F_{\Psi}$ kernel provides the best reconstruction accuracy.

\begin{figure}[t!]
    \centering
    \includegraphics[width=1\linewidth]{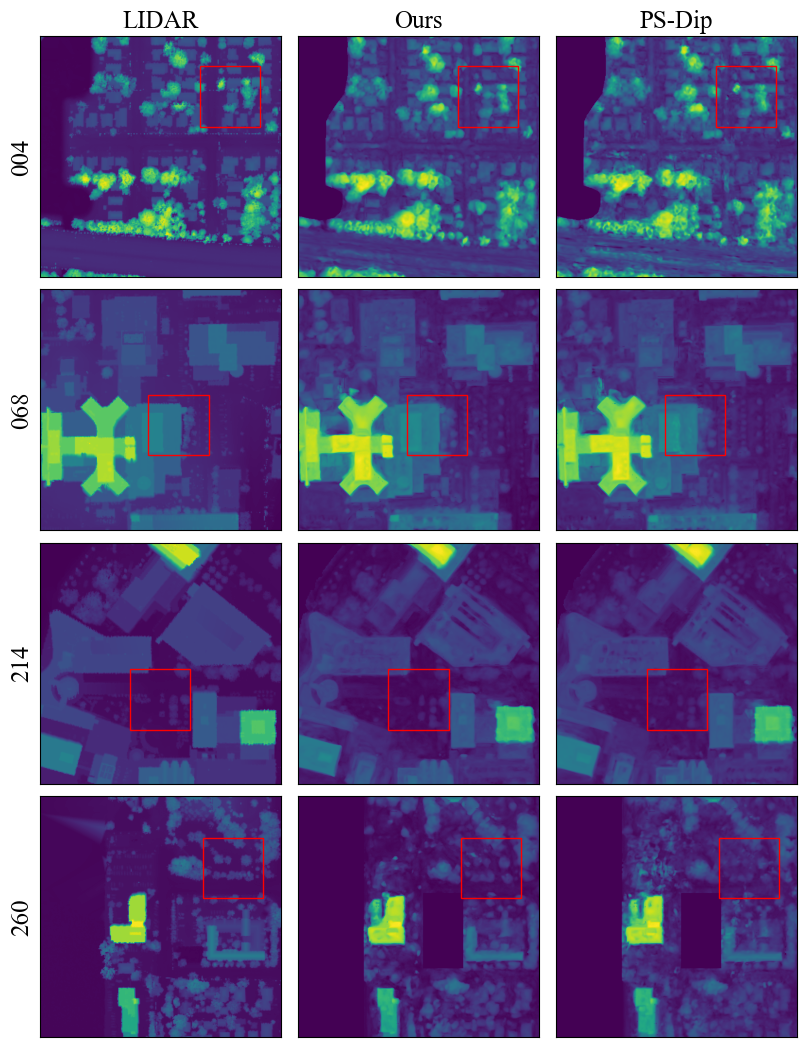}
    \caption{Depth Comparison Results: The results from our method align closer to the LIDAR ground truth, demonstrating the efficacy of our method. Areas highlighted in red are enlarged in the following row.}
    \label{fig:depthcomp1}
\end{figure}

\subsection{Depth Reconstruction Comparisons}
\label{drc}

\begin{figure}[t!]
    \centering
    \includegraphics[width=1\linewidth]{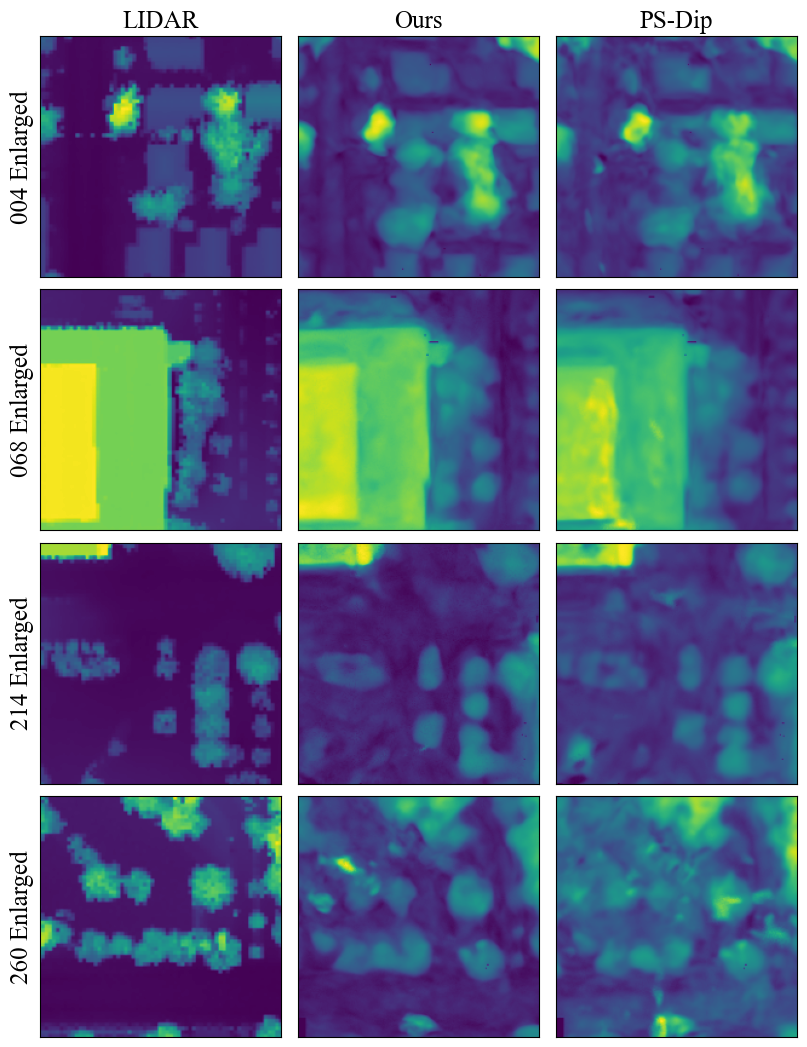}
    \caption{Depth Comparison Results: Areas in red in Figure \ref{fig:depthcomp1} are enlarged to show increased detail in areas with no objects and in the clarity of building edges and rendering of small objects.}
    \label{fig:depthcomp2}
\end{figure}

Four evaluation regions, each covering approximately $0.06$ km$^{2}$, were chosen from WorldView-3 satellite acquisitions available through the 2019 IEEE GRSS Data Fusion Contest (JAX)~\cite{bosch2019semantic, le20192019}. Each region contains between 11 and 24 views, with between 2 and 5 images withheld for evaluation. A slight contrast adjustment was applied but no color correction or normalization was performed. An external bundle adjustment~\cite{mari2021generic} method was used. All images were rendered with the multispectral modal embedding. To measure the quality of constructed site representation, depth maps are generated and compared to LIDAR scans of the scene. The average error is measured with an exclusion mask applied for areas of water. 

\begin{table}[b!]
\centering
\caption{Depth Reconstruction MAE Results: Best result for experiment bolded. On average, our method is able to outperform the best publicly available satellite neural rendering method.}
\begin{tabular}{c|cccc|c|}
\cline{2-6}
\multicolumn{1}{l|}{} & \multicolumn{1}{l}{004} & \multicolumn{1}{l}{068} & \multicolumn{1}{l}{214} & \multicolumn{1}{l|}{260} & \multicolumn{1}{l|}{Mean} \\ \hline
\multicolumn{1}{|c|}{
\rule{0pt}{4ex}
\begin{tabular}[c]{@{}c@{}}Sat-NeRF\\ (Reported)\end{tabular}} & 2.64 & \textbf{1.39} & {\ul 2.38} & 2.37 & 2.195 \\ 
\multicolumn{1}{|c|}{
\rule{0pt}{4ex}\begin{tabular}[c]{@{}c@{}}FusionRF\\ (No $F_{\Psi}$)\end{tabular}} & 1.516 & 1.547 & 2.65 & 2.31 & 2.006 \\ 
\multicolumn{1}{|c|}{\rule{0pt}{4ex}\begin{tabular}[c]{@{}c@{}}FusionRF \\ (No panchromatic images)\end{tabular}} & 1.763 & 1.771 & 2.792 & 2.912 & 2.309 \\ 
\multicolumn{1}{|c|}{\rule{0pt}{4ex}\begin{tabular}[c]{@{}c@{}}FusionRF\\ (Random position $F_{\Psi}$)\end{tabular}} & {\ul 1.494} & {\ul 1.450} & 2.444 & 2.306 & 1.923 \\ 
\multicolumn{1}{|c|}{\rule{0pt}{4ex}\begin{tabular}[c]{@{}c@{}}FusionRF\\ (w/ Modal Embeddings)\end{tabular}} & 1.550 & 1.522 & 2.390 & {\ul 2.041} & 1.876 \\
\multicolumn{1}{|c|}{\rule{0pt}{4ex}\begin{tabular}[c]{@{}c@{}}FusionRF\\ (w/ Full Embedding Strategy)\end{tabular}} & \textbf{1.404} & 1.533 & \textbf{2.350} & \textbf{2.007} & \textbf{1.823} \\ \hline
\end{tabular}%d
\label{tab:depthresults}
\end{table}

In Figures \ref{fig:depthcomp1} and \ref{fig:depthcomp2}, we show a comparison between the generated depth from FusionRF trained with cross-resolution kernel $F_\Psi$ on panchromatic and multispectral data, and that of FusionRF trained with pansharpened data from PSDip~\cite{10744593}. We show visually improved clarity in detailed regions, such as small trees and building edges. Additionally, we observe better performance in reconstructing the ground, with less noise and false objects present. Figure \ref{fig:sharppancomp} shows a visual comparison between depth reconstruction results for a sample scene.

\begin{figure}[t!]
    \centering
    \includegraphics[width=1\linewidth]{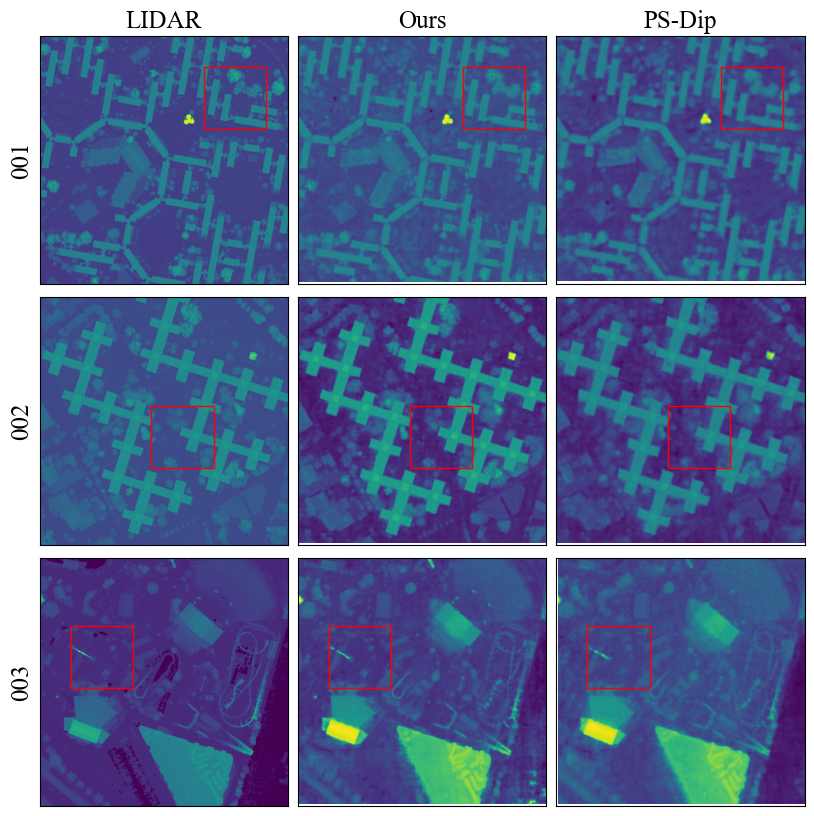}
    \caption{Depth Comparison Results: The results from FusionRF+EO-NeRF optimized over multispectral and panchromatic imagery aligns closer to the LIDAR ground truth than the same model optimized over pansharpened imagery generated by PSDip, demonstrating the efficacy of our method. Areas highlighted in red are enlarged in the following row.}
    \label{fig:eodepth}
\end{figure}

In Table \ref{tab:depthresults}, we show the Mean Absolute Error (MAE) calculated between the LIDAR scan and the generated depth from various ablations of FusionRF. Additionally, we include results from a baseline method, Sat-NeRF~\cite{mari2022sat}. We show that FusionRF is able to reduce reconstruction error by 17\% compared to Sat-NeRF, the performance of FusionRF increases when sparse cross-resolution kernel $F_{\Psi}$ is enabled, the inclusion of both $F_\Psi$ and input panchromatic images is necessary to perform intrinsic pansharpening, and the increase in performance provided by image and modal embeddings. Figures \ref{fig:depthcomp1} and \ref{fig:depthcomp2} visually show depth comparison results for sample scenes in each dataset.

\subsection{Extension to EO-NeRF}
\label{eonerf}

\begin{table*}[b]
\centering
\caption{Comparison of EO-NeRF and FusionRF methods across JAX and IARPA datasets. Best mean result bolded.}
\begin{tabular}{c|cccc|ccc|c|}
\cline{2-9}
\multicolumn{1}{l|}{} & \multicolumn{4}{c|}{\textbf{JAX}} & \multicolumn{3}{c|}{\textbf{IARPA}} & \multicolumn{1}{l|}{} \\
\multicolumn{1}{l|}{} & \multicolumn{1}{l}{004} & \multicolumn{1}{l}{068} & \multicolumn{1}{l}{214} & \multicolumn{1}{l|}{260} & \multicolumn{1}{l}{001} & \multicolumn{1}{l}{002} & \multicolumn{1}{l|}{003} & \multicolumn{1}{l|}{\textbf{Mean}} \\ \hline
\multicolumn{1}{|c|}{
\rule{0pt}{4ex}
\begin{tabular}[c]{@{}c@{}}EO-NeRF\\ (Reported)\end{tabular}} & 1.72 & 1.08 & 1.58 & 1.53 & 2.07 & 2.39 & 1.99 & 1.76 \\
\multicolumn{1}{|c|}{
\rule{0pt}{4ex}
\begin{tabular}[c]{@{}c@{}}FusionRF + EO-NeRF\\ (PSDip Pansharpened)\end{tabular}} & 1.36  & 1.36  & 1.88  & 1.62  & 1.68  & 2.17  & 2.04  & 1.73 \\
\multicolumn{1}{|c|}{
\rule{0pt}{4ex}
\begin{tabular}[c]{@{}c@{}}FusionRF + EO-NeRF\\ (Raw Imagery)\end{tabular}} & 1.29 & 1.15 & 1.72 & 1.41 & 1.51 & 1.70 & 2.29 & \textbf{1.58} \\ \hline
\end{tabular}
\label{tab:eocomparison}
\end{table*}

In order to demonstrate the flexibility of our embedding strategy and cross-resolution kernel, we adapt them to the state-of-the-art method EO-NeRF~\cite{qu2023sat}. As an ablation model, we train EO-NeRF on images produced by the state-of-the-art pansharpening algorithm PSDip~\cite{10744593} with our novel embedding strategy. For this comparison, the original RPCs provided with the images were used for all models. 

\begin{figure}[t]
    \centering
    \includegraphics[width=1\linewidth]{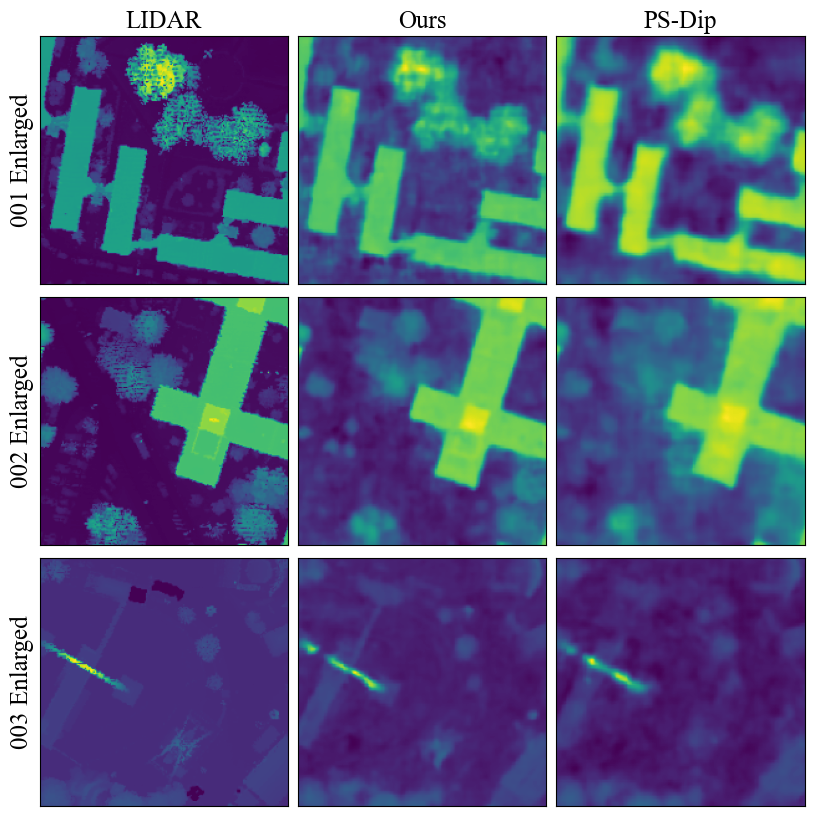}
    \caption{Depth Comparison Results: Areas in red in Figure \ref{fig:eodepth} are enlarged to show increased detail in the clarity of building edges and rendering of small objects.}
    \label{fig:eodepthzoom}
\end{figure}

For additional scenes, three evaluation regions from the 2016 IARPA Multi-View Stereo 3D Mapping Challenge (IARPA)~\cite{8010543} are adopted, each covering $0.06$ km$^2$. Each area contains between 20 and 24 distinct views with 2 to 5 views withheld for validation. A slight contrast adjustment was added but no color correction or normalization was performed. We provide a visual comparison between LIDAR scans of the scene and the depth reconstruction results of our adapted model and the ablation model in Figures \ref{fig:eodepth} and \ref{fig:eodepthzoom}. 

Our comparison shows the increased clarity in depth reconstruction results when using the original multispectral and panchromatic images instead of pansharpening the images upstream of the reconstruction model. We note that the depth reconstruction is clearer in sections of fine detail such as building edges, trees, and small structures.

We calculate the mean absolute error of the adapted and ablation models against the LIDAR scans and report the results in Table \ref{tab:eocomparison}. Additionally, we include the reported EO-NeRF performance on pansharpened imagery. Our results show a 10.23\% decrease in reconstruction error across all datasets, indicating a performance benefit from optimizing over raw multispectral and panchromatic imagery. This adapted model demonstrates that the proposed cross-resolution kernel and embedding strategies can be adapted to any NeRF model that operates on satellite imagery to avoid the pansharpening stage. 

\subsection{Timings}

Using the cross-resolution kernel means that the model must also render an additional eight neighboring support rays for each pixel in the input images. Additionally, FusionRF must render twice the number of pixels as pansharpened methods, since it optimizes over multispectral and panchromatic images independently. This results in a much longer training time than other models. However, since minimal parameters are used in the cross-resolution kernel the overall parameter count is very similar to the baseline models. Additionally, the inference speed of our models are identical to the baseline models since the cross-resolution kernel is disabled during inference. The exact timings and parameter counts are shown in Table \ref{tab:timings}. 

\begin{figure}[t!]
    \centering
    \includegraphics[width=1\linewidth]{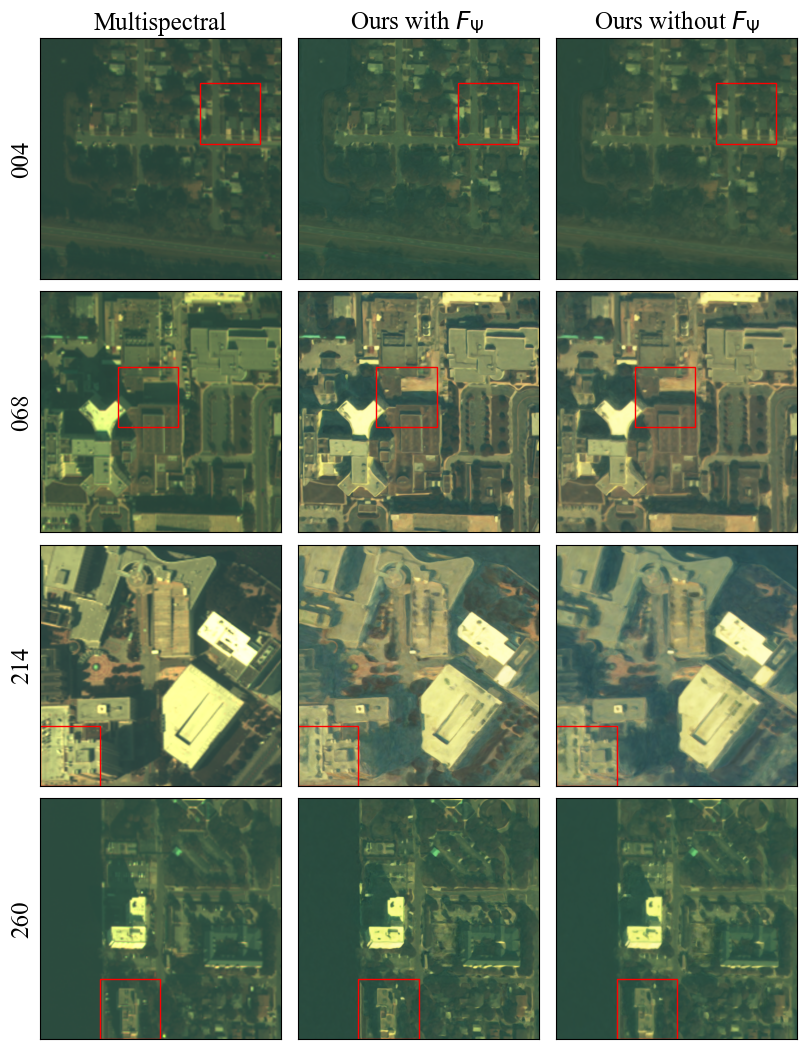}
    \caption{Sharpness Comparison Results: The results from the inclusion of the static cross-resolution kernel show a clear enhancement in perceptual quality and sharpness compared to both the original multispectral image and an ablation of the model with no static cross-resolution kernel. Test images are shown. Areas highlighted in red are enlarged in the following row.}
    \label{fig:sharpcomp1}
\end{figure}

\begin{table}[b!]
\caption{Model Performance: While training takes substantially longer, the inference speed and model parameter count are very near that of the baseline models.}
\label{tab:timings}
\centering
\resizebox{\columnwidth}{!}{%
\begin{tabular}{|c|ccc|}
\hline
\multicolumn{1}{|l|}{}& \textbf{\begin{tabular}[c]{@{}c@{}}Training Time\\ (per epoch)\end{tabular}} & \textbf{\begin{tabular}[c]{@{}c@{}}Number of \\ Trainable Parameters\end{tabular}} & \textbf{\begin{tabular}[c]{@{}c@{}}Inference Speed\\ (per image)\end{tabular}} \\ \hline
\textbf{Sat-NeRF}& $\sim$1h 18min& 2,523,634  & $\sim$41s \\
\textbf{FusionRF}& $\sim$7h 56min& 2,541,117  & $\sim$41s \\
\textbf{EO-NeRF}& $\sim$39min & 781,283 & $\sim$ 35s \\
\textbf{FusionRF + EO-NeRF} & $\sim$6h 18min& 799,030 & $\sim$35s \\ \hline
\end{tabular}%
}
\end{table}

\begin{figure}[t]
    \centering
    \includegraphics[width=1\linewidth]{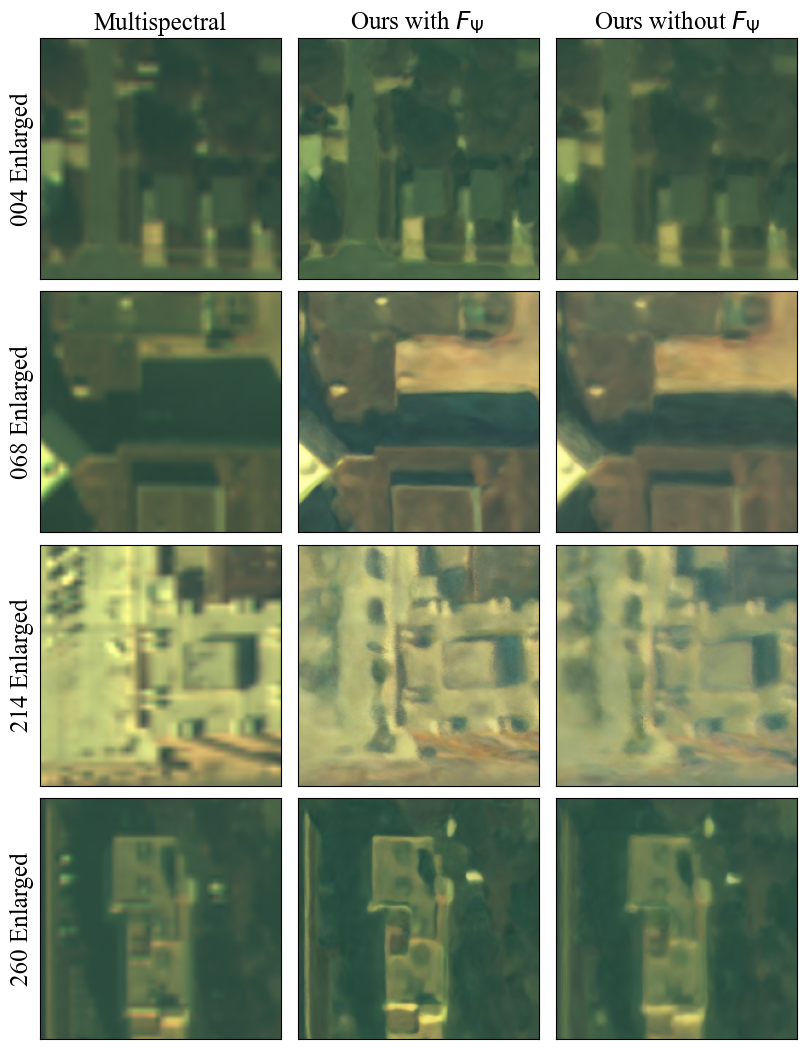}
    \caption{Sharpness Comparison Results: Areas in red in Figure \ref{fig:sharpcomp1} are enlarged to show increased detail in small objects and overall increased sharpness of the scene. Test images are shown.}
    \label{fig:sharpcomp2}
\end{figure}

\subsection{Existing Neural Image Restoration Methods}

In Table \ref{tab:depthresults}, we show that incorporating a more traditional DeBlur-NeRF kernel (FusionRF with Random Position $F_\Psi$) is outperformed by the sparse cross-resolution kernel $F_\Psi$ as proposed, decreasing error by 5\%. As the size of our datasets is vastly reduced compared to those used for DeBlur-NeRF and PDRF, we design a lightweight cross-resolution kernel with a low parameter count to specifically model resolution loss. Our design's effectiveness results in visually sharper and more detailed multispectral images and more accurate depth reconstruction results. 

In Figure \ref{fig:sharpcomp1}, we show an ablation between FusionRF trained with and without $F_\Psi$. Both models were trained with multispectral and panchromatic data, and were generated with the multispectral modality. We note distinct clarity improvements in novel view synthesis shown in Figure \ref{fig:sharpcomp2} provided by training FusionRF with the cross-resolution kernel $F_\Psi$ in both building and ground-level details. Table \ref{tab:depthresults} shows that the inclusion of $F_\Psi$ decreases reconstruction error by 9\%.

\section{Conclusion}
We present FusionRF, an advancement in satellite-based NeRF methods that removes the requirement of pansharpening in satellite image preprocessing. Our model fuses low spatial resolution multispectral and low spectral resolution panchromatic imagery directly from common observation satellites and intrinsically performs pansharpening, rendering sharp novel view images with both high spectral and high spatial resolutions. 

Our method outperforms previous NeRF methods with public codebases on full-channel imagery in digital surface modeling and produces novel view images closer to the original multispectral. This is achieved through novel modal embeddings that allow the model to fuse information between paired panchromatic and multispectral images while a cross-resolution kernel resolves resolution loss in multispectral images. While further work can be done to increase the quality of the neural rendering, our method demonstrates that no optical preprocessing is required to generate high quality scene reconstructions from satellite images.

\section{Acknowledgement}
This research is based upon work supported by the Office of the Director of National Intelligence (ODNI), Intelligence Advanced Research Projects Activity (IARPA), via IARPA R\&D Contract No. 140D0423C0076. The views and conclusions contained herein are those of the authors and should not be interpreted as necessarily representing the official policies or endorsements, either expressed or implied, of the ODNI, IARPA, or the U.S. Government. The U.S. Government is authorized to reproduce and distribute reprints for Governmental purposes notwithstanding any copyright annotation thereon.

\newpage

\begin{IEEEbiography}[{\includegraphics[width=1in,height=1.25in,clip,keepaspectratio]{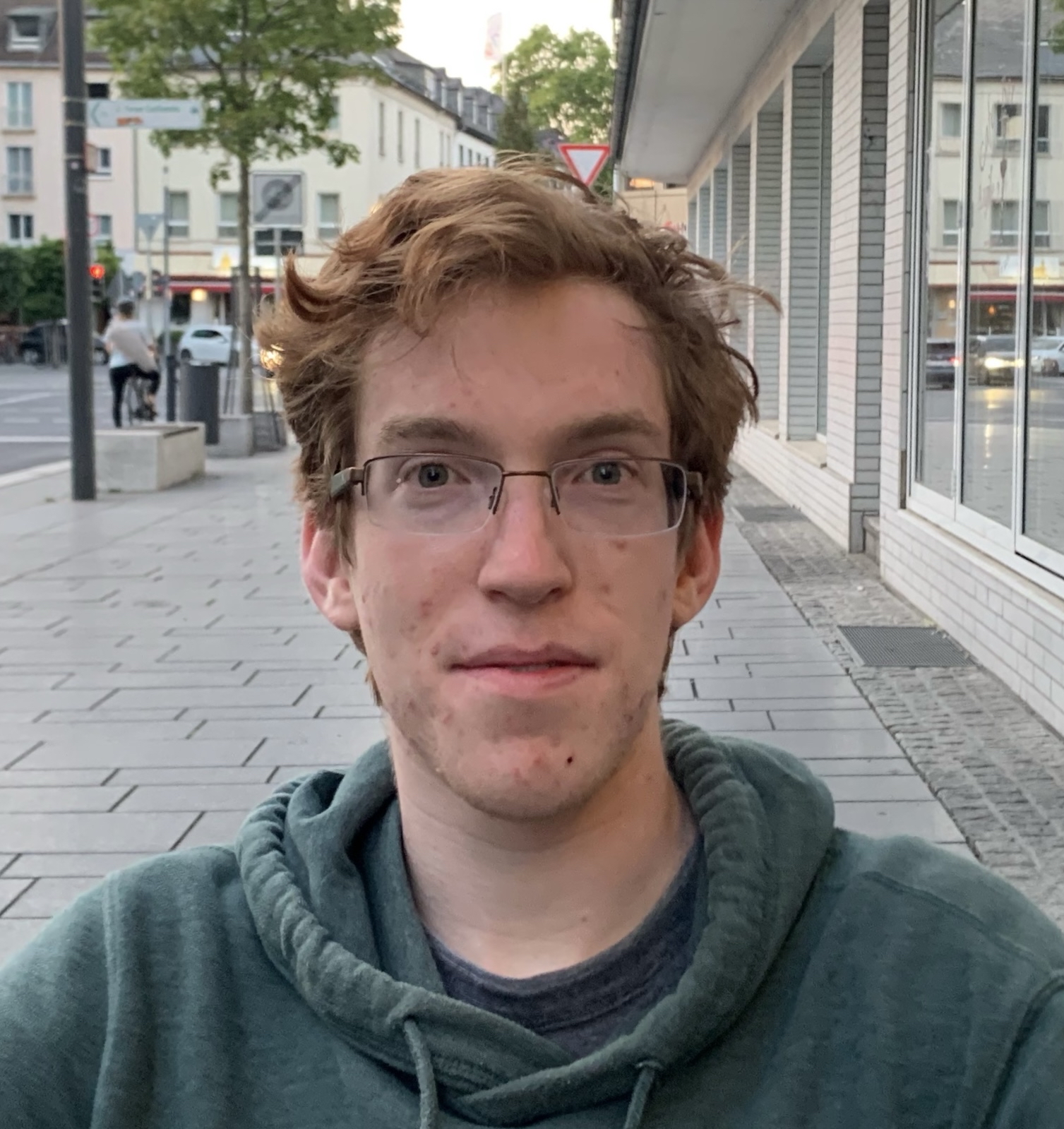}}]{Michael Sprintson}
is a fourth year PhD student at the Johns Hopkins University in the Artificial Intelligence for Engineering and Medicine (AIEM) Lab. He is advised by Prof. Rama Chellappa and Prof. Cheng Peng with research focused on computer vision tasks such as scene understanding and reconstruction. He earned his master's degree in Electrical Engineering from the Johns Hopkins University in 2025 and bachelor's degrees in Electrical Engineering and Computer Science from Rice University in 2022.

\vspace{-4.0em}

\end{IEEEbiography}
\begin{IEEEbiography}[{\includegraphics[width=1in,height=1.25in,clip,keepaspectratio]{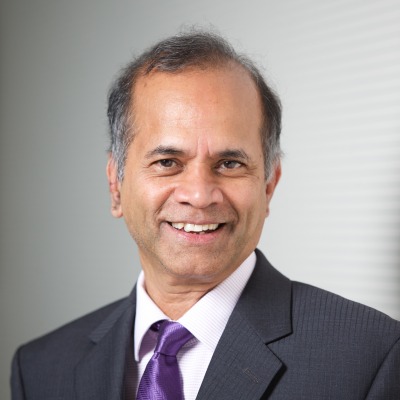}}]{Prof. Rama Chellappa} (Life Fellow, IEEE) is a Bloomberg Distinguished Professor in the Departments of Electrical and Computer Engineering and Biomedical Engineering at Johns Hopkins University (JHU). At JHU, he is affiliated with the Center for Imaging Science (CIS), the Center for Language and Speech Processing (CLSP), the Data Science and AI Institute (DSAI), the Institute for Assured Autonomy (IAA), and MINDS. He also holds a non-tenured appointment as a College Park Professor in the Department of Electrical and Computer Engineering at the University of Maryland. His research interests span computer vision, pattern recognition, machine learning, and artificial intelligence. Chellappa has received numerous honors, including the 2024 Edwin H. Land Medal from Optica (formerly the Optical Society of America), the 2025 Azriel Rosenfeld Lifetime Achievement Award and the 2023 Distinguished Researcher Award from the IEEE Computer Society's PAMI Technical Committee, and the 2020 IEEE Jack S. Kilby Medal for Signal Processing. He is also the recipient of the 2012 K. S. Fu Prize from the International Association of Pattern Recognition, as well as the Society, Technical Achievement, and Meritorious Service Awards from the IEEE Signal Processing Society; the Technical Achievement and Meritorious Service Awards from the IEEE Computer Society; and the Inaugural Leadership Award from the IEEE Biometrics Council. He is an Elected Member of the U.S. National Academy of Engineering and a Foreign Fellow of the Indian National Academy of Engineering. Dr. Chellappa is a Fellow of AAAI, AAAS, ACM, AIMBE, IAPR, IEEE, NAI, Optica, and the Washington Academy of Sciences, and holds nine patents. 
\end{IEEEbiography}
\begin{IEEEbiography}[{\includegraphics[width=1in,height=1.25in,clip,keepaspectratio]{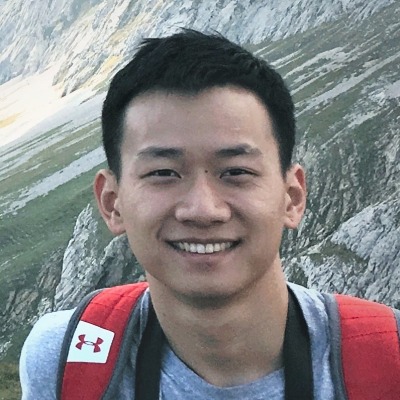}}]{Prof. Cheng Peng} is an Assistant Professor of Data Science at the University of Virginia. Previously, he was a Research Assistant Professor at the Johns Hopkins University and a member of the Artificial Intelligence for Engineering and Medicine (AIEM) Lab. He obtained his PhD in Computer Science from the Johns Hopkins University in 2023 and his master's and bachelor's degrees in Electrical Engineering from the University of Maryland, College Park in 2016 and 2018. Cheng's research focuses on a variety of computer vision tasks in the natural and medical image domains, ranging from low-level vision tasks such as super-resolution, 3D reconstruction, deblurring, etc. to high-level tasks such as segmentation and classification.
\end{IEEEbiography}

\vfill

\end{document}